# Pre-insertion resistors temperature prediction based on improved WOA -SVR

*Honghe Dai[a], Site Mo[a,*], Haoxin Wang[a], Nan Yin[a], Songhai Fan[b,c], Bixiong Li[d]*

[a]College of Electrical Engineering,Sichuan University,Chengdu,610065,Sichuan China
[b]State Grid Sichuan Electric Power Research Institute,Chengdu 610041, Sichuan China
[c]Power Internet of Things Key Laboratory of Sichuan Province, Chengdu 610031, Sichuan China
[d]College of Architecture and Environment,Sichuan University,Chengdu,610065,Sichuan China

ABSTRACT

The pre-insertion resistors (PIR) within high-voltage circuit breakers are critical components and warm up by generating Joule heat when an electric current flows through them. Elevated temperature can lead to temporary closure failure and, in severe cases, the rupture of PIR. To accurately predict the temperature of PIR, this study combines finite element simulation techniques with Support Vector Regression (SVR) optimized by an Improved Whale Optimization Algorithm (IWOA) approach. The IWOA includes Tent mapping, a convergence factor based on the sigmoid function, and the Ornstein-Uhlenbeck variation strategy. The IWOA-SVR model is compared with the SSA-SVR and WOA-SVR. The results reveal that the prediction accuracies of the IWOA-SVR model were 90.2% and 81.5% (above 100°C) in the ±3°C temperature deviation range and 96.3% and 93.4% (above 100°C) in the ±4°C temperature deviation range, surpassing the performance of the comparative models. This research demonstrates the method proposed can realize the online monitoring of the temperature of the PIR, which can effectively prevent thermal faults PIR and provide a basis for the opening and closing of the circuit breaker within a short period.

## 1. Introduction

With the rapid economic development in China, higher demands have been placed on the construction of the power grid, emphasizing the significance of constructing more efficient and robust ultra-high voltage (UHV) and extra-high voltage (EHV) electrical networks. Concurrently, a growing number of switchgear devices, such as Gas-Insulated Switchgears (GIS), Hybrid Gas-Insulated Switchgears (HGIS), and tank circuit breakers, have widely adopted PIR in their designs[1] [2] . During the opening and closing operations of circuit breakers, PIR play a pivotal role in reducing excitation inrush currents and transient overvoltage. Following the operation of opening and closing, PIR accumulate heat due to the passage of electrical current, leading to an increase in their internal temperature. In cases of frequent circuit breaker operations or the passage of substantial currents, excessive temperature rise in the PIR can result in their rupture, preventing the normal execution of subsequent circuit breaker operations and posing safety risks to the power system. If the temperature of the PIR can be monitored online, then stop the tripping or reduce the frequency of tripping when the resistor temperature is too high. However, due to the high-voltage environment in which PIR are situated, direct temperature measurement is unfeasible without the incorporation of sensors. Furthermore, the heating-cooling cycle of PIR constitutes a complex, nonlinear, and strongly coupled process, influenced by a multitude of parameters, rendering it challenging to accurately describe through mathematical models.

Nowadays artificial intelligence has been widely used in engineering prediction, and its classical models include SVR, Extreme Learning Machine (ELM), Long Short-Term Memory (LSTM), Convolutional Neural Network (CNN), et al. For example, Oktay Arikan et al.[3] used an artificial neural network to predict the dielectric parameters of high-voltage cables under overvoltage to assess the short-term dielectric performance of the cables. A. Ajitha et al.[4] proposed a deep learning load forecasting model for the residential sector based on Recurrent Neural Network-LSTM using real-time load data collected from local utility companies. The results obtained show that the prediction accuracy of the proposed model is better than the existing models and proves to be superior. A. A. Milad et al.[5] combined with random forest and multiple Markov chain models to predict asphalt pavement temperature. Ye Zhu et al.[6] develops a model based on the integration of dual attention LSTM and autoregressive moving average (ARMA) to predict aluminum cell temperature. Ephrem Chemali et al.[7] use of deep feedforward networks for battery SOC estimation. Challa Santhi Durganjali et al.[8] used several models to predict PV cell performance. Among them, Kernel Ridge Regression (KRR) performs the best and gives faster (within seconds), valid, and error-free predictions (92.23% accuracy). Ling-Ling Li et al.[9] proposed an aging degree evaluation model based on the WOA for the Optimal ELM algorithm. This study investigated the use of WOA to optimize the input weights and hidden layer bias of ELM to improve its prediction performance. There are also numerous instances of AI applications in situations where direct temperature measurement is challenging. For instance, Boying Liu et al.[10] combined an improved cuckoo search algorithm with an ELM to predict the junction temperature of IGBTs by taking the collector-emitter saturation voltage, collector current, and number of cyclic aging cycles of the IGBTs as inputs. Yuanlong Wang et al.[11] used an Elman neural network (Elman NN) to estimate the temperature variation of lithium-ion batteries in a metal foam thermal management system. Zhang Shuai et al.[12] have employed an enhanced Grey Wolf Algorithm (GWO) to optimize ELM for predicting the coiling temperature of hot-rolled strip steel. Compared to traditional models, their approach exhibits superior accuracy and hit rates. Additionally, Zhou Jianxin et al.[13] have leveraged an improved Pelican Algorithm-optimized LSTM for forecasting the temperature of steel billets in a heating furnace, enabling precise heating, reducing energy consumption, preserving the environment, and enhancing the quality of steel billet production. Han Xiang et al.[14], on the other hand, have utilized a support vector machine optimized by the Enhanced Fish Swarm Algorithm to predict hotspots in transformer windings, effectively enhancing predictive accuracy and achieving favorable forecasting outcomes. Yi Yahui et al. [15] propose a digital twin (DT) technology and LSTM-based method for real-time temperature prediction. The results demonstrate that the proposed real-time temperature prediction framework delivers acceptable accuracy for the constant current discharging and dynamic discharging conditions, which can complete the requirements of practical applications. This study dramatically reduces the response time of temperature prediction and guides optimizing battery thermal management systems (BTMS). J. I. Aizpurua et al.[16] examined the remaining useful life of the transformer under different operating conditions using an extreme gradient boosting (XGB)-based temperature prediction model, and collected uncertainty information from measurements and stochastic processes. Chen[17] et al. developed a predictive model based on Gate Recursive Unit (GRU) that utilizes temperature, fuel, and air time series to predict furnace temperatures.

---

* *Corresponding author.*
  E-mail address: mosite@126.com(S. Mo)

Much of the current research on PIR can be divided into the following categories: power testing, switching techniques, applications in other scenarios, and fault analysis. H. Heiermeier et al.[18] provide alternative test methods consisting of multiple steps for power testing of PIR without compromising its important parameters. In order to mitigate capacitor bank switching transients, synchronized switches using PIR technology have been developed and used in ultra-high voltage transmission systems[19] . They also investigated the coupling relationship between several key factors affecting switching transients; closing target angle, resistor insertion time, and resistor size are all key factors in switching transients for both single-group and back-to-back switches. Kunal A. Bhatt et al.[20] minimize the level of asymmetric DC component of charging current during energization of a shunt reactor using a Controlled Switching Device (CSD) with a PIR-Circuit Breaker, by optimizing the insertion instant, value and Electrical Insertion Time of PIR. T. Eliyan et al.[21] used the PIR of a conventional circuit breaker with three other link technologies to reduce operational overvoltage in wind farms. It was found that PIR and R-L smart choke successfully reduced the operational overvoltage by an average of 52% in wind farms with radial and star topologies. Xiaohui Chen et al. [22] used PIR and CSD for AC filter circuit breakers for transient calculations and gave the optimal configuration. Kunal A. Bhatt et al. [23] proposed using CSD with PIR- Circuit Breaker for further mitigation of the switching surge during energization/re-energization of Uncompensated Transmission Line (UCTL) and Shunt reactor compensated Transmission Line (SCTL). Based on two PIR failures in series and parallel structures, Bo Niu et al. [24] [22] propose methods such as dynamic resistance fitting, acoustic and vibration signal detection, insulation performance test, and heat capacity test during operation. H. Shi et al.[25] simulated three types of faults, short circuit, breakdown,and open circuit, in the PIR. It is found that when the PIR is faulted, the parameters such as harmonic content in the closing inrush, time of occurrence of peak inrush, and three-phase inconsistency are significantly different from the normal condition, and these are applied to the PIR fault diagnosis technique.

This paper presents an IWOA-optimized SVR model for the prediction of PIR temperatures, incorporating finite element simulation data as samples and integrating the Tent mapping, a convergence factor based on the Sigmoid function, and Ornstein-Uhlenbeck variation strategy. The predictive accuracy of the proposed IWOA-SVR temperature forecasting model is compared with that of the SSA-SVR and WOA-SVR models. The results demonstrate that the proposed IWOA-SVR temperature prediction model exhibits superior accuracy, enabling more precise temperature forecasting for PIR.

## 2. IWOA-SVR algorithm

### 2.1. Support vector regression model

SVR is a widely applied regression algorithm, with its core concept centered around constructing a hyperplane with a margin for regression prediction. Its fundamental principle involves minimizing the distance between the hyperplane and the farthest data points while neglecting the loss incurred by data points within the margin, considering only those data points whose absolute difference from the true values exceeds a tolerance $\varepsilon$. To address nonlinear regression problems, SVR utilizes kernel functions to map low-dimensional data into higher-dimensional spaces, effectively approximating the nonlinear problem as a linear regression one. The objective function and constraints [26] are represented as shown in Equation (1):

$$\begin{cases} \min_{\omega,\xi_i,\xi_i^*} \frac{1}{2}\|\omega\|_2^2 + C\sum_{i=1}^{m}(\xi_i + \xi_i^*) \\ y_i - \omega\phi(x_i) - b \leq \varepsilon + \xi_i \quad i=1,2,\ldots,m \\ -(y_i - \omega\phi(x_i) - b) \leq \varepsilon + \xi_i^* \qquad \xi_i \geq 0, \xi_i^* \geq 0 \end{cases} \quad (1)$$

where $C$ is penalty factor. $\xi_i$ and $\xi_i^*$ are slack variables.

### 2.2. WOA

WOA is an intelligent optimization algorithm that simulates the hunting behavior of humpback whales [27] . During the hunting process, humpback whales gather to encircle their prey and subsequently spiral upwards while releasing numerous bubbles to form a bubble net, corralling and driving the prey toward the center of the bubble net before finally consuming it. The core of the WOA algorithm lies in mathematically modeling the humpback whale's actions of encircling, bubble net attack, and prey searching [28] .

1)Encircling the Prey: In the encircling phase, the best whale is defined as the one closest to the prey. Subsequently, other whales update their positions based on the location of the best whale to achieve the encirclement of the prey. The equations for this process are expressed as (2) and (3):

$$x(\lambda+1) = x'(\lambda) - A \cdot |2r_2 \cdot x'(\lambda) - x(\lambda)| \quad (2)$$

$$A = 2a \cdot r_1 - a \quad (3)$$

where $\lambda$ is iteration. $x'(\lambda)$ is the position of best whale; $x(\lambda)$ is the current position of the whale. $A$ is coefficient variable. $a$ is the convergence factor, and will gradually decrease linearly from 2 to 0 as the and will gradually decrease as the increase of $\lambda$、$r_1$、$r_2$ are random numbers between [0,1].

2)Bubble Net Attack: Upon entering the bubble net attack phase, humpback whales not only tighten their encircling formation but also ascend in a spiral motion toward the prey. Assuming that the probabilities of selecting these two behaviors are equal, the updated positions of the whales are computed based on Equation (4).

$$x(\lambda+1) = \begin{cases} x'(\lambda) - A \cdot |2r_2 \cdot x'(\lambda) - x(\lambda)|, P < 0.5 \\ x'(\lambda) + |x'(\lambda) - x(\lambda)| \cdot e^{bl} \cdot \cos(2\pi l), P \geq 0.5 \end{cases} \quad (4)$$

where $b$ is a logarithmic spiral shape constant. $l$ is a random number between [-1,1].

3)Searching for Prey: During the prey searching phase, whales randomly select one whale from the population to approach, thereby enhancing the global search capability of the whale population [29]. The mathematical expression corresponding to this process is given in Equation (5).

$$x(\lambda+1) = x_{rand}(\lambda) - A \cdot |2r_2 \cdot x_{rand}(\lambda) - x(\lambda)| \quad (5)$$

where $x_{rand}(\lambda)$ is a random selection of the location of the whale to be approached.

### 2.3. Improvement of the WOA algorithm

#### 2.3.1. Tent mapping

One of the improvement strategies for swarm intelligence algorithms is to achieve a more uniform initial population distribution, thereby mitigating the effects of uneven distributions caused by random initialization [30]. WOA often employs random initialization of whale positions, which can lead to difficulties in exploring the search space during the initial stages of the algorithm and may result in the algorithm getting stuck in local optima. To address the limitations of random population initialization, the use of chaotic algorithms can promote a more uniform population distribution throughout the search space. Compared to another prevalent chaotic mapping, the logistic map, the tent map exhibits superior uniform traversal characteristics [31]. Consequently, the tent map is employed for initializing the whale population positions. The tent mapping formula is as follows:

$$x_{\lambda+1} = \begin{cases} 2x_\lambda, 0 \leq x_\lambda \leq 0.5 \\ 2(1-x_\lambda), 0.5 \leq x_\lambda \end{cases} \quad (6)$$

where $x_\lambda$、$x_{\lambda+1}$ are the positions of the initial whale after $\lambda$、$\lambda+1$ iterations, respectively.

#### 2.3.2. Convergence factor based on sigmoid function

In the standard WOA, the convergence factor "$a$" linearly decreases from 2 to 0 without considering the need for a thorough search in the early iterations to escape local optima and rapid iterations in the later stages to reduce iteration time. Based on the sigmoid function, a new convergence factor is constructed, expressed as follows:

$$a = 4 \cdot \left( \frac{1}{1+e^{6(\lambda/\lambda_{\max}-1)}} - 0.5 \right) \quad (7)$$

where $\lambda_{\max}$ is the maximum number of iterations.

The improved convergence factor curve is illustrated in Fig.1. In the early iterations, the slope of this curve is relatively low, which is advantageous for allowing the whales in the algorithm to explore the global optimum in a divergent manner. In the later iterations, the improved convergence factor exhibits a steep slope, enhancing the convergence speed. Compared to the original convergence factor, the improved convergence factor performs better in seeking the global optimum and achieves a faster convergence rate.

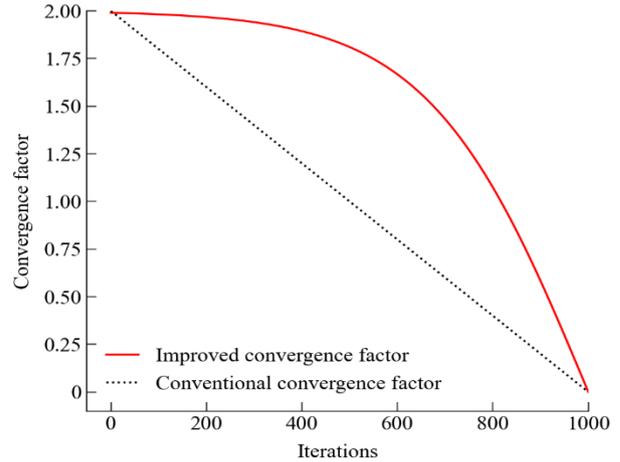

Fig. 1 Improved convergence factor

#### 2.3.3. Incorporating the Ornstein-Uhlenbeck mutation strategy

Traditional mutation choices include the normal distribution or Cauchy distribution[32] [33]. However, when using these distributions, as the probabilities are positive, the optimal values can only mutate in a positive direction, which does not fulfill the requirement for negative-direction mutations. The Ornstein-Uhlenbeck process (OU process), while maintaining randomness, exhibits both positive and negative values [34] [35], and it also possesses a certain degree of temporal correlation. An illustration of one such stochastic process is shown in Fig.2. From Fig.2, it can be observed that the OU process produces values that can be both positive and negative, falling within the range of [-0.24, 0.24], making it a suitable choice for perturbations.

$$X^{t+1} = X_{best} \cdot (1 + OU[B]) \quad (8)$$

$$B = t * m * \lfloor C_1 / (t_{\max} * N) \rfloor \quad (9)$$

where $OU$ is expressed as the value of the Ornstein-Uhlenbeck process. $C_1$ denotes the number of points taken by OU in the horizontal coordinate. $N$ indicates the number of whale populations. $m$ indicates that a position update is being performed for the mth whale. $X_{best}$ is Current optimal fitness. $X^{t+1}$ is the post mutation fitness.

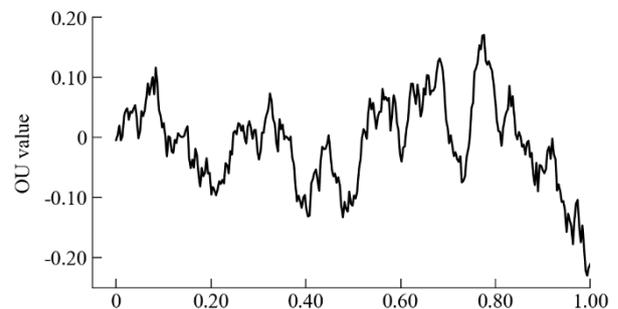

Fig.2 Ornstein-Uhlenbeck stochastic process

While the aforementioned mutations enhance the algorithm's capability to escape local spaces, they do not guarantee that the fitness value of the new position after perturbation mutation will be superior to that of the original position. Consequently, following perturbation mutation updates, a greedy rule is introduced to determine whether the position should be updated by comparing the fitness values of the old and new positions. The greedy rule is defined as follows, where $f(x)$ represents the fitness value of the position.

$$X'_{best} = \begin{cases} X^{t+1}, f(X^{t+1}) < f(X_{best}) \\ X_{best}, f(X^{t+1}) \geq f(X_{best}) \end{cases} \quad (10)$$

where $X'_{best}$ is the more appropriate fitness after comparison.

### 2.3.4. IWOA Implementation Steps

The steps of IWOA incorporating the above improvements are as follows:
**Step.1** Initialize the fundamental parameters of the WOA.
**Step.2** Initialize the whale position using the tent chaotic mapping.
**Step.3** Calculate the fitness of all individual whales and sort them.
**Step.4** Record the position of the best-performing whale in terms of fitness.
**Step.5** Apply the Ornstein-Uhlenbeck mutation to the best-performing whale's current position, generating a mutated solution, and calculate the fitness of the mutated solution according to Equation (8).
**Step.6** If the fitness of the mutated solution surpasses that of the original, replace the original with the mutated solution.
**Step.7** Compute the magnitude of the convergence factor using Equation (9) and update the position vectors according to Equation (10).
**Step.8** If the algorithm satisfies the termination condition, terminate and output the best solution; otherwise, proceed to **Step.3**.

### 2.4. IWOA performance test

To evaluate the performance of the proposed IWOA, seven benchmark functions from CEC2005 were selected for algorithm comparison. The comparison was conducted in conjunction with the GWO, Sparrow Search Algorithm (SSA), and WOA. The number of whale populations is set to 30, the dimensions of the test functions are 10, 30 and 50, the maximum number of iterations is 1000, and the number of independent runs of each test function is 30. Tab.1 provides information about the test functions, while Tab.2 presents the results of the function tests.

Tab.1 Information of test functions

| Label | Function | Range | Optimal value | Peak value |
|---|---|---|---|---|
| F1 | Sphere | [-100,100] | 0 | single peak |
| F2 | Schwefel2.22 | [-10,10] | 0 | single peak |
| F3 | Schwefel1.2 | [-100,100] | 0 | single peak |
| F4 | Schwefel2.21 | [-100,100] | 0 | single peak |
| F5 | Ackley | [-32,32] | 0 | multi-peak |
| F6 | Penalized1 | [-50,50] | 0 | multi-peak |
| F7 | Penalized2 | [-50,50] | 0 | multi-peak |

The specific results of the 4 algorithms optimizing the 7 benchmark test functions are shown in Tab.2 (all optimal values have been bolded.). In Tab.2, on the single-peak functions F1-F4, the IWOA algorithm has a mean, standard deviation and optimal value of 0. The mean value is the theoretical optimal value, which is a large improvement compared to the remaining three algorithms. While on the multi-peak function F5, IWOA is close to the optimal value of WOA, but slightly inferior to the comparison algorithms in mean and standard deviation. On the multi-peak functions F6 and F7, IWOA is better than the comparison algorithm in terms of mean and standard deviation. The convergence curves for the seven test functions when the dimension of the test function is 30 are shown in Fig.3. From Fig.3, it can be seen that on the F1-F7 functions, the convergence speed of IWOA is greatly improved relative to the comparison algorithm, especially on the F1-F4 functions. The mean, standard deviation, optimal value and convergence speed of the IWOA algorithm in these seven test functions are significantly better than the comparison algorithms in general, which proves the effectiveness of the improvement strategy.

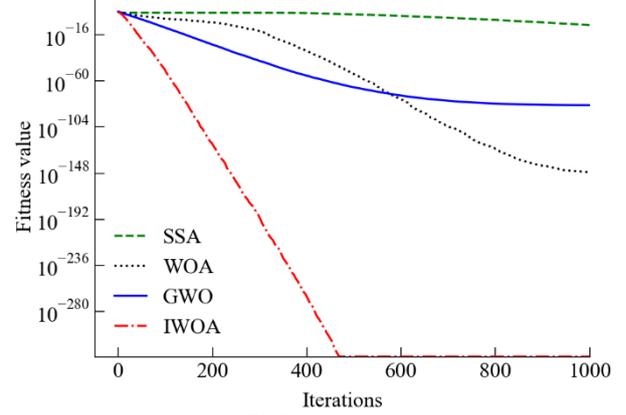
(a)F1 Convergence curve

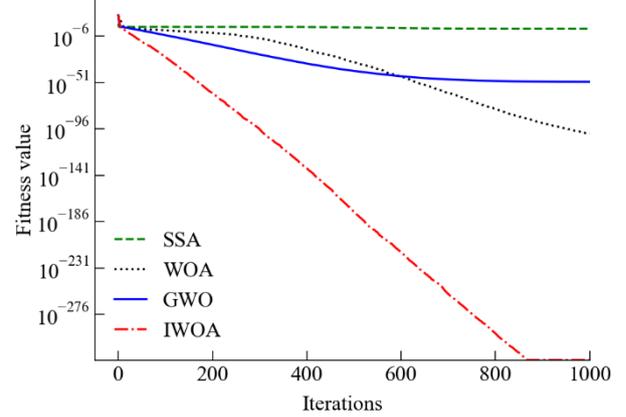
(b)F2 Convergence curve

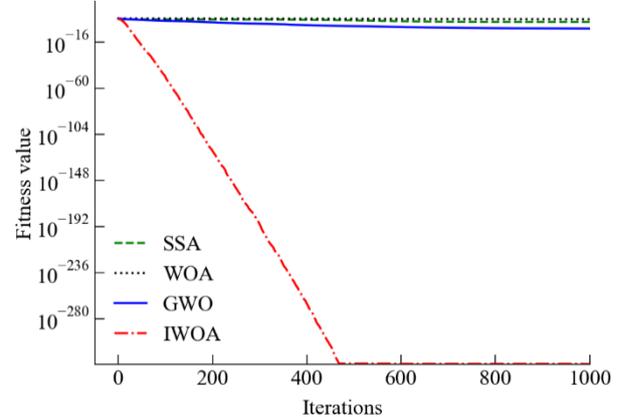
(c)F3 Convergence curve

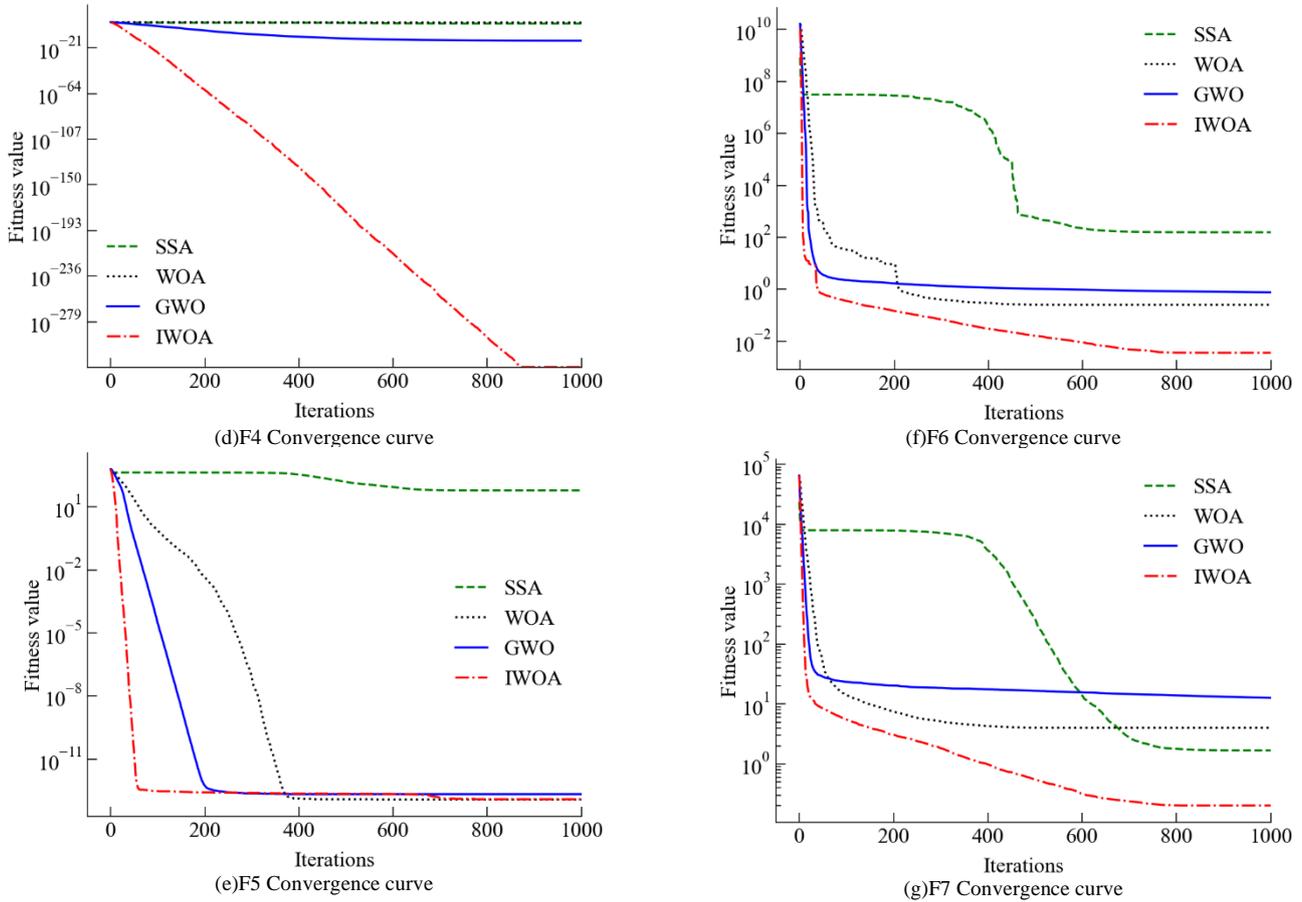

Fig.3 Convergence curve of the test function when dimension=30

Tab.2 Function test results

| Function | Algorithm | dimension=10 | | | dimension=30 | | | dimension=50 | | |
|---|---|---|---|---|---|---|---|---|---|---|
| | | Average value | Standard deviation | Optimal value | Average value | Standard deviation | Optimal value | Average value | Standard deviation | Optimal value |
| F1 | GWO | 3.29E-117 | 1.09E-116 | 5.75E-125 | 2.75E-85 | 8.23E-85 | 3.53E-88 | 1.00E-78 | 1.65E-78 | 1.18E-82 |
| | SSA | 7.09E-10 | 2.20E-10 | 2.56E-10 | 1.02E-08 | 2.07E-09 | 7.21E-09 | 8.89E-08 | 2.84E-08 | 5.40E-08 |
| | WOA | 2.29E-156 | 0.00E+00 | 3.43E-175 | 1.7E-149 | 9.4E-149 | 1.7E-167 | 1.08E-146 | 5.76E-146 | 1.27E-166 |
| | IWOA | 0 | 0 | 0 | 0 | 0 | 0 | 0 | 0 | 0 |
| F2 | GWO | 3.08E-71 | 9.30E-71 | 1.21E-75 | 1.29E-52 | 4.78E-52 | 4.50E-54 | 7.84E-49 | 1.19E-48 | 1.10E-50 |
| | SSA | 4.97E-03 | 2.67E-02 | 3.44E-06 | 5.28E-01 | 7.29E-01 | 7.11E-04 | 3.88E+00 | 1.76E+00 | 3.09E-01 |
| | WOA | 5.75E-105 | 3.09E-104 | 1.85E-118 | 1.6E-102 | 8.5E-102 | 1.6E-113 | 7.46E-102 | 3.90E-101 | 3.09E-115 |
| | IWOA | 0 | 0 | 0 | 0 | 0 | 0 | 0 | 0 | 0 |
| F3 | GWO | 3.11E-27 | 1.49E-26 | 3.57E-38 | 2.65E-05 | 7.58E-05 | 9.99E-11 | 1.69E-01 | 3.41E-01 | 1.59E-06 |
| | SSA | 1.98E-09 | 1.14E-09 | 6.77E-10 | 4.89E+01 | 4.95E+01 | 7.85E+00 | 5.27E+03 | 2.52E+03 | 1.71E+03 |
| | WOA | 7.54E+00 | 1.70E+01 | 1.29E-15 | 1.95E+04 | 1.17E+04 | 1.54E+03 | 1.30E+05 | 3.32E+04 | 4.57E+04 |
| | IWOA | 0 | 0 | 0 | 0 | 0 | 0 | 0 | 0 | 0 |
| F4 | GWO | 2.78E-28 | 1.45E-27 | 1.77E-33 | 2.71E-16 | 5.90E-16 | 2.09E-19 | 4.61E-14 | 6.31E-14 | 3.29E-16 |
| | SSA | 1.49E-05 | 4.25E-06 | 7.92E-06 | 3.33E+00 | 1.25E+00 | 9.89E-01 | 1.74E+01 | 3.33E+00 | 9.72E+00 |
| | WOA | 6.47E-01 | 2.40E+00 | 1.51E-07 | 3.41E+01 | 3.34E+01 | 1.83E-02 | 6.75E+01 | 2.57E+01 | 1.88E-01 |
| | IWOA | 0 | 0 | 0 | 0 | 0 | 0 | 0 | 0 | 0 |
| F5 | GWO | 5.06E-15 | 1.63E-15 | 4.00E-15 | **7.08E-15** | 1.54E-15 | 4.00E-15 | 7.67E-15 | **1.12E-15** | 4.00E-15 |

| Function | Algorithm | dimension=10 | | | dimension=30 | | | dimension=50 | | |
|---|---|---|---|---|---|---|---|---|---|---|
| | | Average value | Standard deviation | Optimal value | Average value | Standard deviation | Optimal value | Average value | Standard deviation | Optimal value |
| | SSA | 7.69E-01 | 9.41E-01 | 7.11E-06 | 2.01E+00 | 7.44E-01 | 2.69E-05 | 3.09E+00 | 6.88E-01 | 1.88E+00 |
| | WOA | 3.17E-15 | 1.76E-15 | 4.44E-16 | 3.88E-15 | **2.55E-15** | 4.44E-16 | **3.17E-15** | 2.19E-15 | 4.44E-16 |
| | IWOA | 4.47E-15 | 3.01E-15 | **4.44E-16** | 4.00E-15 | 2.95E-15 | **4.44E-16** | 5.06E-15 | 2.92E-15 | **4.44E-16** |
| F6 | GWO | 8.19E-03 | 1.34E-02 | 1.47E-07 | 2.53E-02 | 2.05E-02 | 6.57E-03 | 6.37E-02 | 8.83E-02 | 1.48E-02 |
| | SSA | 2.11E-01 | 5.68E-01 | **2.13E-12** | 5.21E+00 | 3.68E+00 | 5.85E-01 | 9.21E+00 | 3.90E+00 | 4.68E+00 |
| | WOA | 1.66E-04 | 2.02E-04 | 1.63E-05 | 8.44E-03 | 6.24E-03 | 8.60E-04 | 1.29E-02 | 9.28E-03 | 4.56E-03 |
| | IWOA | **7.84E-06** | **1.15E-05** | 8.94E-07 | **1.19E-04** | **6.08E-05** | **3.96E-05** | **6.36E-04** | **2.13E-04** | **3.14E-04** |
| F7 | GWO | 1.45E-02 | 3.37E-02 | 1.77E-07 | 4.24E-01 | 2.43E-01 | 8.49E-02 | 1.29E+00 | 4.18E-01 | 4.89E-01 |
| | SSA | 1.10E-03 | 3.30E-03 | **3.79E-12** | 5.63E-02 | 2.76E-01 | **7.44E-10** | 3.96E+01 | 7.45E+01 | 5.75E-02 |
| | WOA | 1.66E-03 | 3.91E-03 | 1.75E-05 | 1.33E-01 | 9.79E-02 | 9.96E-03 | 5.44E-01 | 3.29E-01 | 8.61E-02 |
| | IWOA | **2.53E-05** | **6.41E-05** | 1.60E-06 | **6.73E-03** | **6.17E-03** | 6.15E-04 | **3.00E-02** | **1.93E-02** | 9.69E-03 |

## 3. Simulation and Predictive Evaluation of PIR Temperature Rise

### 3.1. PIR temperature rise simulation

When a high current passes through the PIR, it accumulates a significant amount of heat. The PIR, which is at a temperature higher than the ambient temperature, undergoes heat conduction, convection, and thermal radiation interactions with its surroundings, including nearby objects and the environment. The entire physical field involves three main aspects: the electrical field, the heat field, and the flow field. The interrelationships among these three physical fields are depicted in Fig.4.

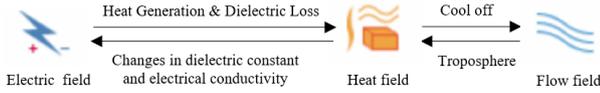

Fig.4 Relationship between multiple physical fields

In the COMSOL Multiphysics environment, a two-dimensional axisymmetric simulation model (Fig.5) is constructed. The model comprises 35 PIR sheets, two copper plates, an insulating rod, a tank filled with $SF_6$, and an outer steel casing [36] . Since the PIR sheets are tightened with bolts, no gap is explicitly modeled between them in the simulation. However, there is a necessary gap between the insulating rod and the PIR sheets due to their mating relationship, which affects the cooling of the PIR and cannot be neglected. Therefore, this gap should be included in the simulation, approximately set to 1 mm. The current is injected into the bottom copper plate, while the topmost copper plate is grounded. Fig.5 also illustrates the temperature field distribution across the PIR after simulation. Tab.3 shows the physical parameters of the main substances of the simulation model. The parameters of the PIR are taken from the technical documentation provided by the manufacturer and the parameters of the insulated pole are taken from literatures[37] [38] , where the PIR conductivity is shown in Fig.6:

Tab.3 physical parameters of the main substances

| Physical parameter | PIR | Insulated pole | $SF_6$ |
|---|---|---|---|
| Relative permittivity | 5 | —— | —— |
| Thermal conductivity/[W/(m·K)] | 4 | 0.5 | $\lambda_g$ |
| Constant pressure heat capacity /[J/(kg·K)] | 890 | 789.52 | $C_p$ |
| Conductivity /[S/m] | $\rho$ | —— | —— |
| Density /[kg/m³] | 2250 | 2290 | 30 |
| Kinetic viscosity/[Pa·s] | —— | —— | $\mu$ |

$$\lambda_g = 4.37e^{-3} - 5.78e^{-5}T + 4.79e^{-7}T^2 - 9.19e^{-10}T^3 + 8.18e^{-13}T^4 - 2.82e^{-16}T^5 \tag{11}$$

$$C_p = -218.4 + 4.73T + 7.50e^{-3}T^2 + 5.67e^{-6}T^3 - 1.66e^{-9}T^4 \tag{12}$$

$$\mu = 2.88e^{-7} + 5.51e^{-8}T - 1.68e^{-11}T^2 + 1.39e^{-15}T^3 \tag{13}$$

where $T$、$\lambda_g$、$C_p$、$\mu$ are the temperature, thermal conductivity, constant pressure heat capacity, kinetic viscosity of $SF_6$.

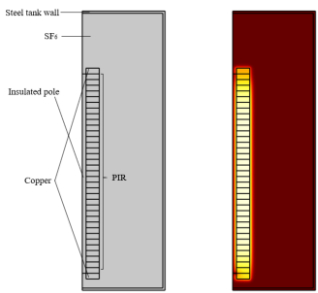

Fig.5 PIR simulation model

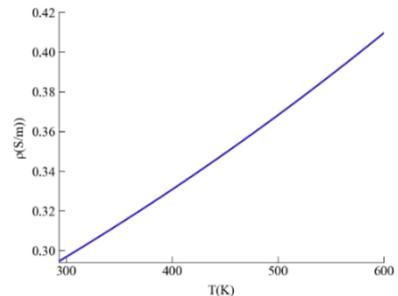

Fig.6 PIR conductivity

7
COMSOL utilizes finite element analysis, and thus, the mesh used significantly impacts the computational results. A denser mesh can lead to a large computational load, while a coarser mesh may result in less accurate results. Three different grid sizes are selected, i.e., each PIR sheet is divided into 10*10, 15*15, and 20*20 cells. The calculation results for different grids are shown in Fig.7. The results show that the change temperature is negligible when the number of grids is raised from 100 to 400. In order to reduce the amount of calculation, 100 grids per PIR sheets are used in this paper. The final mesh for the model is depicted in Fig.8.

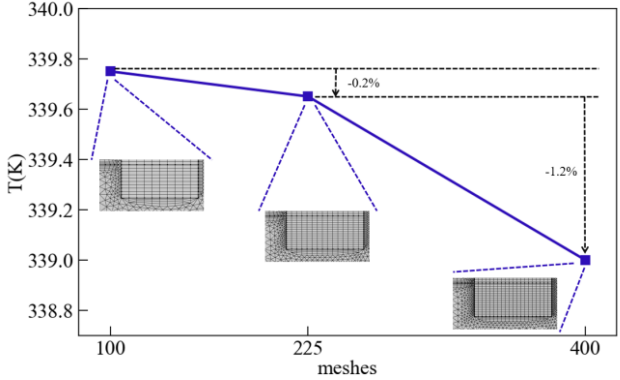

Fig.7 Temperature of different mesh

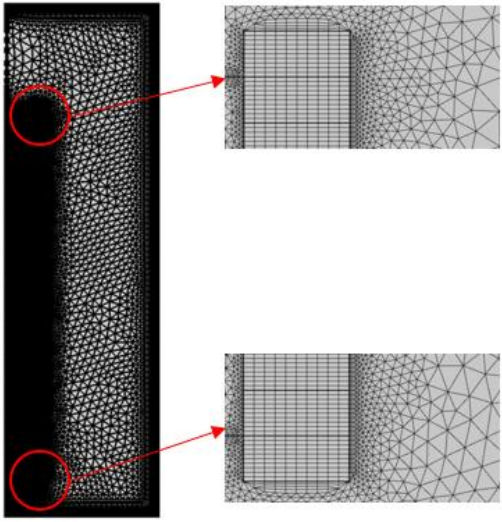

Fig.8 Model grid segmentation

### 3.2. Data Set Acquisition

There are five factors that affect the temperature of the PIR after cooling: the value of the closing current, the closing phase angle, the input current time, the cooling time, and the initial temperature of the PIR. The closing current is approximated as a sinusoidal wave. 500kV circuit breaker PIR sheets for thermal capacity experiments into the larger 1600A sinusoidal current. PIR input time is generally 8-12ms. In order to increase the diversity of the dataset, the ranges of current and input current time were set to [0A,1600A] and [7ms,12ms], respectively. In accordance with State Grid requirements, for important lines, when single-phase reclosing is unsuccessful and three-phase tripping occurs, the trial delivery time is not less than 30 minutes from the last tripping time interval, so the maximum prediction time is set to 1800s. For the prediction of the closing phase angle is set to [0,2π]. The PIR sheets require that the temperature rise should not exceed 125°C at an ambient temperature of 25°C. The minimum initial temperature is set to 293K (ambient temperature is 20°C), and the maximum temperature is set to 393K, which is close to the maximum allowable temperature of the resistor (423K). The other factors are ignored as they hardly change during the switching process. The ranges for the above five factors were set as Tab.4. In finite element simulation, since the mesh cannot be encrypted indefinitely, sometimes there are singular values, which may simulate inaccurate maximum values. The maximum temperature of the PIR has an extremely important role in its state evaluation. After many experiments, the temperature distribution of the PIR is more uniform after cooling (Fig.5), i.e., the maximum value of the PIR is very close to the average value. Therefore, the simulation and prediction of the temperature value of the PIR in this paper are the average temperature of 35 PIR sheets.

Tab.4 The range of factors affecting temperature

| Variable | | Range |
|---|---|---|
| Current | $I$/A | [0,1600] |
| Input current time | $t_1$/ms | [7,12] |
| Cooling time | $t_2$/s | [0,1800] |
| Closing phase angle | $\omega$ | [0,6.28] |
| Initial temperature | $T_o$/K | [293,393] |

A total of 4,000 sets of variables are randomly generated using tent mapping. A joint simulation using MATLAB and COMSOL is conducted to obtain the post-cooling temperatures of the PIR under 4000 different scenarios. Some of the simulation data are shown in Tab.5.

Tab.5 Partial simulation data

| $I$/A | $t_1$/ms | $t_2$/s | $\omega$ | $T_o$/K | Simulation temperature |
|---|---|---|---|---|---|
| 500.33 | 10.82 | 1259.97 | 0.49 | 331.63 | 355.62 |
| 1159.43 | 7.80 | 744.43 | 5.86 | 378.46 | 380.10 |
| 448.62 | 11.30 | 1210.33 | 2.33 | 340.16 | 361.60 |
| 801.76 | 10.05 | 701.00 | 0.76 | 383.74 | 386.02 |
| 1142.43 | 9.26 | 700.13 | 1.01 | 351.90 | 355.55 |
| 1357.53 | 9.56 | 1189.12 | 0.32 | 323.35 | 344.62 |
| 715.39 | 9.20 | 436.06 | 0.99 | 327.53 | 328.64 |
| 1300.23 | 11.49 | 644.20 | 4.65 | 343.90 | 349.36 |
| 1497.72 | 10.07 | 1016.34 | 3.62 | 350.78 | 363.48 |

### 3.3. Analysis of projected results

To validate the predictive performance of the IWOA for optimizing SVR, a testing dataset consisting of 30% of the data is selected. Two other models, WOA-SVR and SSA-SVR, are chosen for comparison. Evaluation of the models is conducted using three metrics: the coefficient of determination ($R^2$), mean squared error (MSE), and mean absolute error (MAE). The results of the models are summarized in Tab.6, and the hit rate of each model is presented in Fig.9 and Fig.10.

$$R^2 = 1 - \frac{\sum_{i=1}^{n}(y_i - y_i^{'})^2}{\sum_{i=1}^{n}(y_i - \bar{y})^2} \tag{14}$$

$$MSE = \frac{1}{n}\sum_{i=1}^{n}(y_i^{'} - y_i)^2 \tag{15}$$

$$MAE = \frac{1}{n}\sum_{i=1}^{n}\left|y_i^{'} - y_i\right| \tag{16}$$

where $y_i$ is actual temperature. $y_i'$ is predicted temperature. $\overline{y}$ is actual temperature average. $n$ is predicted sample size.

Tab.6 Comparison of prediction results

| Model | Norms | | |
|---|---|---|---|
| | $R^2$ | MSE | MAE |
| SSA-SVR | 0.99242 | 5.59811 | 1.87360 |
| WOA-SVR | 0.99286 | 5.26979 | 1.82626 |
| IWOA-SVR | **0.99371** | **4.64762** | **1.76508** |

In Tab.6 (all best values have been bolded), it is evident that the proposed IWOA-SVR model outperforms the other two models. It achieves reductions in both MSE and MAE, while simultaneously increasing $R^2$. In Fig.9, the horizontal axis represents the magnitude of the difference between the simulated and predicted values, and the vertical axis represents the proportion of the number of samples in the interval of the difference to the total samples, i.e., the hit rate. It can be seen that IWOA-SVR is lower than SSA-SVR and WOA-SVR by 1.1% and 1% in the interval [-1,1], respectively. But in the other three intervals, it is IWOA-SVR that has the highest hit rate. Especially in the interval [-3,3], the hit rate of IWOA-SVR is 90.2%, which is 3% higher than that of SSA-SVR. The hit rate of IWOA-SVR is 96.3% in [-4,4], which is 1.6% (SSA-SVR) and 1.5% (WOA-SVR) ahead of the other models, respectively.

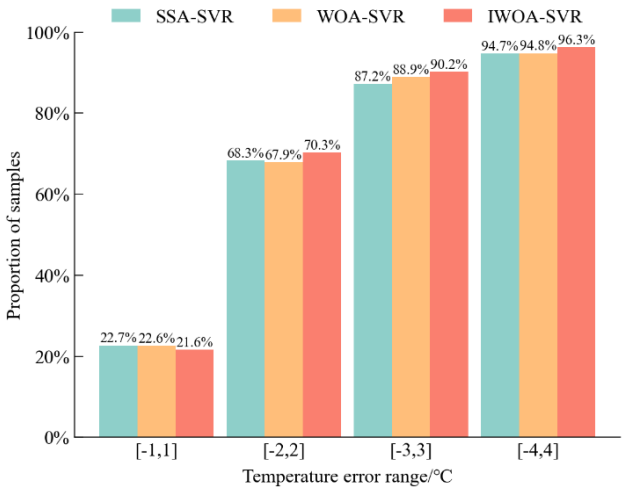

Fig.9 Hit ratio

The prediction accuracy of the model in high temperature environments is critical for the operation of the PIR at high temperatures. The PIR sheets require that the temperature rise should not exceed 125°C at an ambient temperature of 25°C. Fig.10 illustrates the difference between the simulated temperature and the model-predicted temperature at simulated temperatures above 100°C. In the [-1,1] error interval, the hit rate of IWOA-SVR is slightly lower than the other two. Whereas, in the other three error intervals, IWOA-SVR has the highest hit rate. In particular, in the [-4,4] interval, which is more concerned with practical engineering, the hit rate of IWOA-SVR is 6% (SSA-SVR) and 3.8% (WOA-SVR) ahead of the other models, respectively. These results demonstrate the validity of the model and prove the high feasibility of the proposed method for high-precision temperature prediction of PIR in the real engineering field.

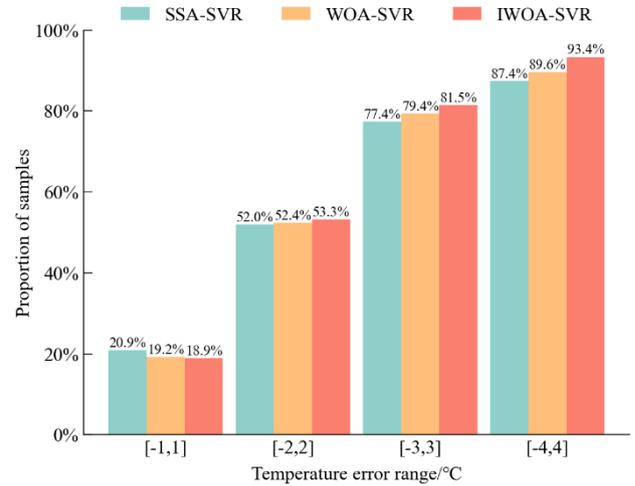

Fig.10 Hit ratio above 100°C

## 4. Conclusion

This study has made improvements to the traditional WOA in terms of initial population, convergence factor, and mutation algorithm. The IWOA-SVR has been applied to the prediction of the PIR temperature, leading to the following key findings:

1) When testing the IWOA against WOA, GWO, and SSA using seven test functions, IWOA outperforms the other three algorithms in terms of the mean, standard deviation, and best value for six of the functions. Moreover, IWOA exhibits a significant advantage in convergence speed across all seven test functions relative to the other three algorithms. This suggests that the IWOA excels in convergence speed and accuracy compared to three other algorithms, and it is less prone to getting stuck in local optima. Therefore, IWOA is more advantageous in finding the optimal penalization factor and the hyperparameters of RBF kernel function in SVR.

2) IWOA-SVR is applied to the temperature prediction of the PIR sheets. Compared with WOA-SVR and SSA-SVR, the results show that the $R^2$, MSE, and MAE of IWOA-SVR are enhanced in temperature prediction. The hit ratio of IWOA-SVR in predicting the temperature is 90.2% within ±3°C, and 96.3% within ±4°C. For temperatures above 100°C, the hit ratio is 81.5% within ±3°C and 93.4% within ±4°C, which is a significant improvement relative to the other two models.

3) Most of the current researches focus on the power testing, switching technology, application, and fault judgment of PIR. The method proposed in this paper can realize the online monitoring of the temperature of the PIR, which can avoid thermal failure of PIR and provide reliable evidence for the opening and closing of the circuit breaker in the power grid in a short period of time.

**Funding**

This research was supported by Science and Technology Project of State Grid Corporation of China (Research and application of fusion perception, intelligent diagnosis and service enhancement technologies at the end of rural power network, No. 52199922000M).